\documentclass[10pt,conference,a4paper]{IEEEtran}
\usepackage[utf8]{inputenc}

\usepackage{times}

\usepackage{graphicx}
\usepackage{subfigure}
\DeclareGraphicsExtensions{.png,.eps,.ps,.pdf}

\usepackage{url}
\usepackage[english]{babel}

\usepackage{amsmath}
\usepackage[linesnumbered,lined,boxed,commentsnumbered]{algorithm2e}
\usepackage{listings}
\usepackage{graphicx}
\usepackage{siunitx}
\usepackage{dirtytalk}
\usepackage{verbatim}
\usepackage{listings}
\usepackage{color}
\usepackage{xcolor}
\usepackage{caption}
\usepackage{subcaption}

\definecolor{dkgreen}{rgb}{0,0.6,0}
\definecolor{gray}{rgb}{0.5,0.5,0.5}
\definecolor{mauve}{rgb}{0.58,0,0.82}

\RestyleAlgo{ruled}
\SetKwComment{Comment}{/* }{ */}
\SetKw{KwBy}{by}

\begin{document}
\title{Identification and Anonymization of Named Entities in Unstructured Information Sources for Use in Social Engineering Detection}

%%% Authors
\author{\IEEEauthorblockN{Carlos Jimeno Miguel}
	\IEEEauthorblockA{Public University of Navarre\\
		Pamplona, Spain\\
		carlos.jimeno@unavarra.es}
	\and
	\IEEEauthorblockN{Raúl Orduna Urrutia}
	\IEEEauthorblockA{Public University of Navarre // Vicomtech\\
		Pamplona, Spain // San Sebastián, Spain\\
		raul.orduna@unavarra.es // rorduna@vicomtech.org}
	\and
	\IEEEauthorblockN{Francesco Zola}
	\IEEEauthorblockA{Vicomtech\\
		San Sebastián, Spain\\
		fzola@vicomtech.org}}

\maketitle

\begin{abstract}
	This study addresses the challenge of creating datasets for cybercrime analysis while complying with the requirements of regulations such as the General Data Protection
	Regulation (GDPR) and Organic Law 10/1995 of the Penal Code. To this end, a system is proposed for collecting information from the \textit{Telegram} platform,
	including text, audio, and images; the implementation of speech-to-text transcription models incorporating signal enhancement techniques; and the
	evaluation of different Named Entity Recognition (NER) solutions, including \textit{Microsoft Presidio} and AI models designed using a \textit{transformer}-based architecture.
	Experimental results indicate that \textit{Parakeet} achieves the best performance in audio transcription, while the proposed NER
	solutions achieve the highest \textit{f1-score} values in detecting sensitive information. In addition, anonymization metrics are presented that
	allow evaluation of the preservation of structural coherence in the data, while simultaneously guaranteeing the protection of personal information and
	supporting cybersecurity research within the current legal framework.
\end{abstract}

\begin{IEEEkeywords}
	Telegram, crawling, dark web, crime, organized crime, audio, anonymization, NER
\end{IEEEkeywords}
{\bf Contribution type:}  {\it Original research}

\section{Introduction}
The democratization of cybercrime enabled by \textit{Crime-as-a-Service} (CaaS) has generated an exponential increase in the frequency and sophistication of attacks. In this context, the \textit{SafeHorizon} project emerges as a European initiative aimed at addressing this threat by developing tools for monitoring digital criminal activities. Monitoring these platforms involves collecting and processing large volumes of data from diverse sources; however, this activity raises critical challenges from the perspective of personal data protection (under regulatory frameworks such as the General Data Protection Regulation, GDPR) \cite{UE2016Reglamento679} and the storage of content whose possession is considered criminal under norms such as Organic Law 10/1995 of the Penal Code \cite{ley_organica_10_1995}. This tension between data protection and analytical utility is one of the central challenges of the project: excessively aggressive anonymization destroys the context and relationships between entities that characterize CaaS networks, making it impossible to detect social engineering patterns, while insufficient anonymization violates the GDPR and could allow the re-identification of individuals.

This work establishes replicable methodologies and tools to operate under the above constraints, with the objective of extracting relationship structures characteristic of fraudulent activities from the analysis of unstructured content present on platforms commonly used by criminals. To this end, information from multiple
formats---text, audio, and images---will be processed and unified in order to identify the underlying conceptual structure linking actors, attack methods, technical infrastructures, and transaction flows. These findings will serve as input for a subsequent analysis of such frauds and their validation by the law enforcement agencies of member states.

To achieve this objective, the following points will be developed: 1) Research into methods and tools for collecting non-traditional intelligence sources frequented by criminal actors, identifying named entities in those sources, and anonymizing said entities. 2) Implementation of the researched tools and re-training of models from the literature to improve the current methods used by the technology center. 3) A comparative evaluation of the performance of the above contributions, selecting the method that yields the best results for each stage.

\section{State of the Art}
In the current context, multiple approaches exist for removing or protecting sensitive information in documents, known as \textit{Privacy-Enhancing Technologies} (PETs). These include techniques such as pseudonymization, data masking, encryption, generalization, and differential privacy, among others. Anonymization, the central subject of this work, constitutes one of the most robust techniques within this set, as it seeks the irreversible elimination of personally identifiable data, thereby fulfilling the most stringent requirements of the GDPR.

Early approaches to data anonymization present statistical models such as \textit{k-anonymity} (P. Samarati and L. Sweeney, 1998) \cite{samarati1998protecting}, \textit{l-diversity} (A. Machanavajjhala, \textit{et al.}, 2007) \cite{LdiversityPaper}, \textit{t-closeness} (N. Li, \textit{et al.}, 2007) \cite{tclosenessPaper}, and \textit{$\delta$-presence} (Nergiz \textit{et al.}, 2007) \cite{nergiz2007hiding}, which describe the identifiability property---the capacity to infer the parameters of statistical models from a series of observations---of anonymous tabular data. This evolution reveals that excessive anonymization could lead to a loss of context, which may constitute a contradiction in the task of generating relational patterns of social engineering.

Transferring these mechanisms to a context where information is not presented in tabular form requires the recognition, in unstructured sources, of entities classifiable as sensitive information. Such techniques involve searching for known \textit{substrings} and regular expressions. The problem with these classical methods is their limited ability to scale to new entity types that may be considered sensitive information in future revisions of existing regulations. For this reason, machine learning methods have been gaining ground over time.

Articles such as those by D. Asimopoulos et al. (2024) \cite{BenchmarkModels} \cite{AsimoEfficacyAI} present comparative studies of different artificial intelligence architectures for entity extraction from plain text. Most publications suggest custom implementations of pre-trained models fine-tuned on domain-specific datasets. Table \ref{tab:comparativa_propuestas} lists several of the published proposals.
\begin{table}[htb]
	\centering
	\caption{Implementations proposed in the state of the art}
	\label{tab:comparativa_propuestas}
	\begin{tabular}{l l l l}
		\hline
		\hline
		Article & Author & Model or Tool\\
		\hline
		\cite{AsimoEfficacyAI} & D. Asimopoulos, \textit{et al.} & CRF/LSTM/ELMo/Transformer\\
		\cite{Hassan2018NER} & F. Hassan, \textit{et al.} & CRF\\
		\cite{Hassan2019AutoAnon} & F. Hassan, \textit{et al.} & Skip-gram\\
		\cite{Siniosoglou2024TextAnon} & I. Siniosoglou, \textit{et al.} & CRF/LSTM/ELMo\\
		\cite{balancingWithPresidio} & S. Patchipala & Microsoft Presidio\\
		\hline
		\hline
	\end{tabular}
\end{table}

This article will consider the alternatives of a model based on the \textit{Transformer} architecture, due to its high performance compared to other models in natural language processing tasks, and the \textit{Microsoft Presidio} tool, given its ease of use, good results, low processing time, and high customizability to the requirements of the problem at hand.

The existence of multimedia messages of different types on social networks such as \textit{Telegram} generates the need to adapt these techniques to the collection of sources such as images and audio. For images, the technology center has a textual description tool and an optical character recognition (OCR) mechanism that will be integrated into the general project workflow. For audio, the performance of different transcription models will be compared by applying signal enhancements to audio recordings made without professional quality, which is common in the cybercrime domain. For this purpose, a voice enhancement and noise reduction system is proposed, based on studies such as that by A. Aben \textit{et al.} \cite{Aben2025WhisperAccessibility}.

\section{Evaluation Methods}
This research work presents an analysis to define a method consisting of obtaining information from unstructured sources, processed to provide relevant data
for an ongoing investigation or operation (intelligence), entity identification, and anonymization of the analyzed results. To
achieve this workflow, a methodological study of each of the aforementioned phases will be conducted.

\subsection{Obtaining Intelligence from Unstructured Sources}
In this phase, the objective is to add a data collector for public/private \textit{Telegram} channels and groups and an audio message transcription model, with signal preprocessing to enhance voice, to the technologies provided by the project partners. These technologies include a \textit{darkweb} market crawler, an image description mechanism, and an OCR system.

In the analysis of the transcription model, a comparative study of different pre-trained models will be conducted before and after signal processing. The goal of this addition is to integrate a new information source from which to collect data. To evaluate this comparison, the \textbf{W}ord \textbf{E}rror \textbf{R}ate (\textit{WER}) metric \cite{Marangon2023WER} will be used, a standard within automatic speech recognition models. A weighting will be applied for each type of error committed; for the proposed experiments, the following weights are assigned: 0.10 for insertions and 0.45 for both deletions and substitutions.

\subsection{Identification of Sensitive Entities under Current Legislation}
Currently, a tool is available that uses regular expressions to recognize a series of named entities (names, surnames, postal addresses, among many others) in web pages collected as plain text files.

These expressions require considerable ongoing maintenance to track new types of data considered sensitive in subsequent revisions of the aforementioned legal frameworks. The use of machine learning mechanisms allows this effort to be significantly reduced. Therefore, the study proposed in the methodology consists of comparing the following algorithms:
\begin{itemize}
	\item The entity identification tool based on regular expressions proposed by the project partners.
	\item A customized architecture of the \textit{Microsoft Presidio} recognition module (hereafter \textit{Presidio}).
	\item A \textit{fine-tuned Transformer} model based on a pre-trained base model.
\end{itemize}

The best method will be selected in terms of \textit{f1 score} \cite{Tigerschiold2022AccuracyMetrics}. This metric is extensively used in artificial intelligence models due to its utility for evaluating models in a general manner. In essence, the \textit{f1 score} provides a weighted balance between the model's precision and its recall.

\subsection{Text Anonymization}
\label{metodos-de-evaluacion-anonimizacion-de-textos}
The tool currently used by the technology center is also capable of performing anonymization. Once the relevant entities for the case have been recognized, it proceeds to replace all their occurrences with a \textit{hash}. This same substitution method has been reviewed and improved with an implementation where entities recognized and labeled by the proposed \textit{fine-tuned} model are replaced by a string containing their class and a \textit{hash}.

In this stage, the current method of the center will be compared against the anonymization tool offered by \textit{Presidio} and the implementation carried out in this research for the \textit{fine-tuned} model. Without a robust and parameterizable metric for characteristics of the anonymized output, several metrics will be proposed to enable the evaluation of new more appropriate methods.

To this end, a series of concepts oriented toward evaluating the previous process are presented, along with the mathematical formalizations of the proposed metrics.

\subsubsection*{Information Loss}
Evaluates the amount of original useful information that is not available in the anonymized output. Let:
\begin{itemize}
	\item $X$ be the set of words that make up an input text
	\item $X'$ be the set of words that make up the anonymized version of text $X$
	\item $w : w \in X$ each of the distinct words that make up the input text
	\item $p(w)$ the frequency of occurrence of a word $w$ in text $X$
	\item $E(X)$ the Shannon entropy \cite{KARACA2022231}
\end{itemize}

The following definition of Shannon entropy will be used:
$$ E(X) = -\sum_{w \in X} p(w) \, log_2 \, p(w) $$

To define information loss as follows:
$$ \text{Information loss}(X, X') = E(X) - E(X') $$

Thus, the domain of this last function lies in the range $[-\infty, \infty]$. Where:
$$
\text{Information loss}(x)
\begin{cases}
	> 0 \text{ Information is lost} \\
	= 0 \text{ Information remains intact} \\
	< 0 \text{ New information is generated}
\end{cases}
$$

\subsubsection*{Per-token Consistency (C)}
Evaluates the capacity for an entity to always be anonymized in the same way. Let $T$ be the set of all unique \textit{tokens} recognized in an input; per-token consistency is defined as:
$$ C = \frac{1}{|T|} * \sum_{t \in T} \frac{\text{Correct conversions of \textit{t}}}{\text{Total occurrences of token \textit{t}}} $$

The domain of the function comprises the interval [0, 1]. Where 0 indicates that all \textit{tokens} are anonymized differently each time, and 1 that each unique \textit{token} is always anonymized in the same way.

\subsubsection*{Collision Degree (G)}
Evaluates the capacity of the anonymization process to generate unique and distinct values for each different entity. Let:
\begin{itemize}
	\item $T$ be the set of original \textit{tokens}
	\item $H$ be the set of generated \textit{hashes}
	\item $f: T \rightarrow H$ the function that generates the anonymization \textit{hash}
\end{itemize}

For each \textit{hash} $h \in H$, the set of \textit{tokens} that generate said \textit{hash} is defined as:
$$ G_h = \{ t \in T \, : f(t) = h \} $$

Therefore, the collision degree is defined as:
$$ G = \frac{|\{h \in H \, : |G_h| = 1\}|}{|H|} $$

This metric quantifies the proportion of unique \textit{hashes} (without collisions) with respect to the total number of generated \textit{hashes}. The domain of the function is the interval $[0, 1]$, where:
\begin{itemize}
	\item $G = 1$ indicates a total absence of collisions (each \textit{token} generates a unique \textit{hash})
	\item $G = 0$ indicates that all \textit{hashes} have multiple collisions (no \textit{hash} is unique)
\end{itemize}

\subsubsection*{Error Rate}
Evaluates the proportion of incorrect assignments taking into account consistency and collisions.

Let $\alpha$ be a parameter that determines the weight assigned to each component of the function. The error rate is defined as:

$$ Error = 1 - (\alpha * C + (1 - \alpha) * G) $$\\
The domain of the function comprises the interval [0, 1]. Where 0 indicates that each unique \textit{token} is encoded in the same way without collisions, and 1 that \textit{tokens} are classified differently each time, generating collisions.

\subsubsection*{Average Correlation Preservation}
Serves as a measure of the sensitivity of the function responsible for the anonymization process. Let:
\begin{itemize}
	\item $A$ and $B$ be two different anonymized texts that share \textit{tokens}
	\item $H_A$ and $H_B$ be the sets containing the \textit{hashes} generated for each of the texts
	\item $T$ be the set representing the \textit{tokens} shared by both texts
	\item $f : T \rightarrow H$ the function that generates the anonymization \textit{hash} for the text
\end{itemize}

For each pair of \textit{hashes} derived from the same \textit{token} $t$, the Levenshtein distance (L) \cite{DominguezPrieto2023Levenshtein} is defined as the function that models the number of modifications, insertions, or deletions required to generate the \textit{hash} $f_b(t) \in H_B$ from the \textit{hash} $f_a(t) \in H_A$.

Based on these definitions, average correlation preservation is described as:
$$ \text{Avg correlation} = \frac{1}{|\{H_A \cup H_B\}|} * \sum_{t \in T} 1 - (\frac{L(f_a(t), f_b(t))}{max(f_a(t), f_b(t))}) $$

The domain of the function comprises the interval [0, 1]. Where 0 indicates that the \textit{hashes} are completely different, and 1 that they are identical.

\section{Experimental Framework}
A first prototype is presented, since no prior algorithms or tools existed, which serves as a proof of concept for collecting intelligence from sources without a general search system. Next, the integration of two new entity identification tools will be described: a proposed architecture for the \textit{Microsoft Presidio} recognition module, and a custom \textit{Transformer}-based model. Finally, the integration of a new anonymization engine, based on the \textit{Microsoft Presidio} anonymization module, will be presented, followed by a comparative evaluation using the metrics described in section \ref{metodos-de-evaluacion-anonimizacion-de-textos}, comparing this engine against the existing implementation at the research center and its adapted revision for the \textit{Transformer}-based model developed here.

\subsection{Obtaining Intelligence through Telegram}
Since search engines within social networks such as \textit{Telegram} do not easily return results for channels or groups engaged in illicit activities, alternatives must be sought. Commercial repositories (such as \textit{TGstat.com}) exist where such searches can be performed; however, access to these is restricted by their prohibitive cost for many research contexts, and their lack of documentation on their search mechanisms and algorithms prevents replication. Therefore, a method is presented based on Open Source Intelligence (OSINT) techniques such as \textit{Google dorking} or \textit{hacking}.

For this purpose, the `\textit{site}' clause is used in the search, which allows focusing results within a specific web domain. In the case of \textit{Telegram}, two types of link are available depending on the visibility of the resource to be consulted:
\begin{itemize}
	\item \textit{t.me/s/\textbf{identifier}}: Specifies access to a public resource, under the name \textbf{identifier}, anonymously from a web browser.
	\item \textit{t.me/+\textbf{hash}}: Specifies the invitation link to a private resource. This \textbf{hash} is randomly generated when the invitation is created, making it difficult to trace its origin.
\end{itemize}
Combining both link types with the search term (\textit{Carding} or \textit{Ransomware}, for example), a query can be made to the \textit{Google} API and the title, description, and access link of the results can be retrieved.

Once the collection is complete, messages will be extracted from the resource. In the case of a public resource, simple web \textit{crawling} is performed on the page, extracting the content from the HTML elements corresponding to the messages. For private resources, together with the invitation link and a previously created \textit{Telegram} account, the group is accessed via an implementation built using the \textit{Telethon} library. This library implements a communication interface with the \textit{Telegram} API that allows operations such as joining a group and inspecting its messages. A sequence diagram of the development is presented below.
\begin{figure}[ht]
	\centering{
		\includegraphics[width=6cm]{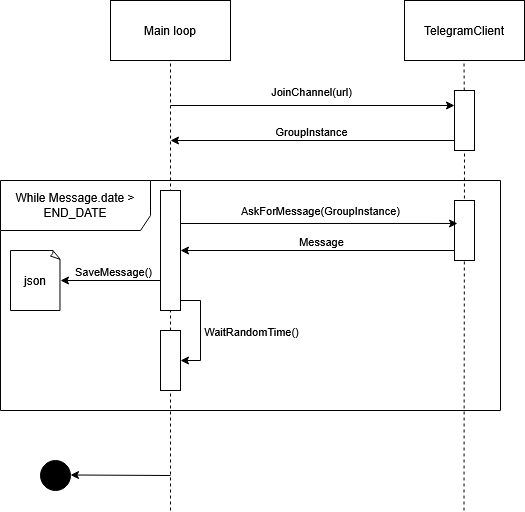}
	}
	\caption{Sequence diagram of message collection}
	\label{fig:sequence-messages}
\end{figure}

All existing messages up to a specified date will be retrieved. Furthermore, it is noted that in order to avoid exceeding the application's usage thresholds, a considerable random delay is introduced for each retrieved message (between 30 and 60 seconds).

This development is transferable to other social networks, always taking into account the need to study the technical and legal feasibility of their integration as an information source, through the analysis of official APIs for their manipulation and the platform's terms of use and service.

\subsection{Obtaining Intelligence from Audio Sources}
For the development and validation of the module, 3000 entries from the \textit{espnet/yodas-granary} \cite{espnet-yodas-granary-ds} dataset (approximately 25 hours of audio) were selected from the \textit{HuggingFace} repository, along with the lightest versions of the main models from the three best-positioned companies within the same repository:
\begin{enumerate}
	\item \textit{Nvidia} with its model \textbf{parakeet-tdt-0.6b-v3}.
	\item \textit{OpenAI} with its model \textbf{whisper-small}.
	\item \textit{Meta} with its model \textbf{wav2vec2-base-960h}.
\end{enumerate}
Additionally, \textbf{pyannote/segmentation-3.0} was used as the Voice Activity Detection (VAD) model. Using this model and the Python \textit{noisereduce} library, an audio signal can be filtered to enhance the voice by considering only the regions of the signal where voice is present.

The steps performed to clean an audio signal were: 1) Convert the audio signal and its sampling rate into a format readable by the VAD model. 2) Recognize the intervals where no voice activity is present. 3) Construct a new ``noise'' signal from the recognized intervals. 4) Filter the audio by passing the original signal, its sampling rate, and the ``noise'' signal from the previous step as arguments to \textit{noisereduce}. 5) Reconstruct the original signal with the filtered noise.

Once this process is implemented, the experiment consists of a \textit{benchmark} of the three aforementioned models in terms of accuracy (evaluating their WER) and execution time (in seconds), loading the 3000 samples, instantiating the models in English, and generating predictions on both the original and preprocessed signals.

\subsection{Entity Identification with Microsoft Presidio}
\label{identificacion-entidades-presidio}
Since \textit{Presidio} allows the integration of classes, methods, and workflows, a three-step architecture is proposed. First, use the regular expressions already present in the center's original method, to leverage existing knowledge. Second, use the general-purpose GLiNER model \cite{zaratiana2023glinergeneralistmodelnamed} with which the tool has native support. Third, configure a pre-trained BERT-type model, through the \textit{HuggingFace} \textit{transformers} library, to extend the previous knowledge to new entity types.

The architecture will be evaluated on the \textit{ildpil/text-anonymization-benchmark} dataset \cite{Pilan2025TAB}, consisting of a set of free-format judicial texts with a large number of entities categorized as sensitive.

Although the structure of the dataset may resemble a real-internet context, a prior data processing step is required. First, only the \textit{text} fields (the original transcription of the full text) and \textit{entity mentions} (a dictionary where each entity contains its class and start/end indices in the text) will be collected. Second, given the length of the texts, they will be segmented so that the maximum of 384 tokens of the selected BERT model is not exceeded, calculating for each segment the offset of the indices of each word it contains. Finally, the dataset labels will be standardized to match those used by \textit{Presidio}.

The following steps will be executed to configure the proposed architecture within the module. For the case of regular expressions, each one will be declared within the \textit{Pattern} class under the desired name and its specific regular expression, as shown in listing \ref{code:presidio-pattern}.
\begin{lstlisting}[language=Python, caption=Regular expression configuration, label=code:presidio-pattern]
	p = Pattern(name="document_number", regex=r"\d+\/\d{2}", score=0.7)
\end{lstlisting}
For the GLiNER and BERT models, they must be declared in the same way, specifying the name of the model to be used from the \textit{HuggingFace} repository (\textit{urchade/gliner multi pii-v1} \cite{zaratiana2023gliner} and \textit{dbmdz/bert-large-cased-finetuned-conll03-english} \cite{Fliegner2023BERTConll03} respectively), a dictionary of how the class names recognized by the model will be standardized to \textit{Presidio} names, and the execution environment to be used (GPU or CPU).

Finally, it is necessary to reconstruct the tool's predictions given that the original dataset format has been altered. To do so, the text is simply reconstructed from the calculated segments and the predicted classes are reverted to the original dataset classes---not the \textit{Presidio} ones---in order to correctly evaluate the tool's performance on free-format texts.

\subsection{Entity Identification with a Transformer-based Model}
First, a feasibility study will be conducted by replicating the article by D. Asimopoulos, \textit{et al}. \cite{BenchmarkModels} using the base model \textit{google/bert-base-cased} \cite{GoogleBertBaseCased2025}, since the implementation of their own \textit{Transformer} model is not specified, under the \textit{CoNLL-2003} dataset \cite{conll-2003-ds}.

It should be noted that for this feasibility phase as well as the following ones, the hyperparameters selected for the model are: \textit{learning rate} of $2 * 10^{-5}$, \textit{Batch size} of 16, \textit{Number of epochs} of 3, and \textit{Weight decay} of 0.1.

Once the feasibility of using this base model has been verified, a first fine-tuning will be performed on the \textit{tner/ontonotes5} dataset \cite{tner-ontonotes5}. Since the dataset contains information related to other fields of study, the inputs will be formatted to obtain only the \textit{tokens} and their labels, discarding the rest. From here, it remains to declare the above hyperparameters, train, and evaluate performance.

To validate the above training, a subset of the same dataset of judicial rulings from the previous section will be used---specifically 200 entries. The objective is to evaluate the performance of the previous \textit{fine-tuning} in a context where texts are not prepared for training. To achieve this, the texts will be segmented so as not to exceed the maximum number of tokens processable by this model (512 in this case), and the inputs will be formatted to conform to the IOB2 format understood by the model. As in section \ref{identificacion-entidades-presidio}, once predictions are made, the texts will be reconstructed to regroup the subdivisions the model has applied to certain words (\textit{subtokens}) back into their original \textit{tokens}, thus correctly evaluating the model's performance on free-format texts.

Finally, the model will be adapted to the objective of obtaining a dataset free of regulatory restrictions for cybercrime analysis. For this purpose, three datasets have been selected (\textit{Josephgflowers/CENSUS-NER-Name-Email-Address-Phone}, \textit{Josephgflowers/PII-NER}, and \textit{ai4privacy/pii-masking-400k}) that present classes corresponding to the cybercrime domain: full names, credit cards, national ID numbers, passports, email addresses, phone numbers, and locations.

The selected data present several issues that must be addressed. There are more classes than are relevant to the study, and entries with invalid values, such as null or empty values, which will be omitted and filtered in a first step. Additionally, class names vary between datasets, so these classes will be homogenized under the following labels: `\textit{ADDRESS}', `\textit{CREDITCARDNUMBER}', `\textit{EMAIL}', `\textit{IDCARDNUM}', `\textit{NAME}', `\textit{PASSPORT}', and `\textit{PHONE}'. It is also necessary to convert the above classes to the IOB2 format understood by the model.
\begin{figure}[ht]
	\centering{
		\includegraphics[width=8.5cm]{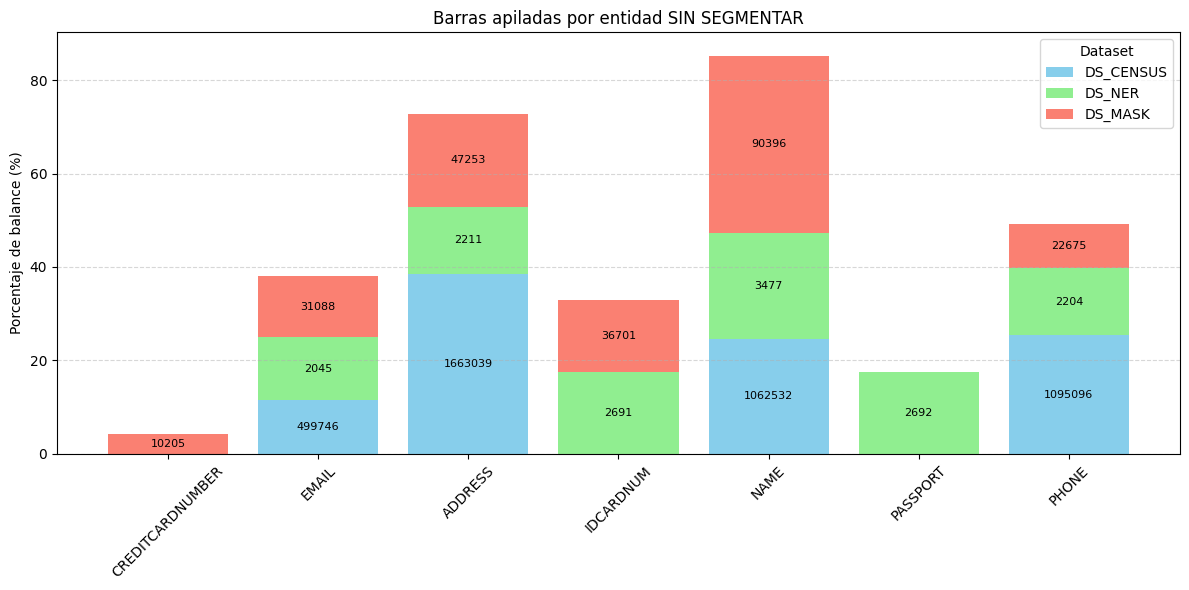}
	}
	\caption{Distribution of the datasets used}
	\label{fig:dataset}
\end{figure}

Even with this processing, a class imbalance is observed, with many entries in classes such as \textit{NAME} and \textit{ADDRESS} and very few in \textit{CREDITCARDNUMBER} or \textit{PASSPORT}. To minimize the impact of this situation on retraining, the density of entries per class will be redistributed for the training and validation sets to obtain the best possible distribution. Moreover, although not implemented due to the project's time constraints, the artificial generation of entries is proposed to append to this dataset via the \textit{Faker} library, including more entities from the underrepresented classes and new classes such as IP addresses or \textit{blockchain} addresses corresponding to the studied domain.

To use this library, it is necessary to initialize the main class, the \textit{Faker()} class, which generates data types with a seed; in the experiment, seed `12345' was used. Next, the \textit{paragraph(NUM\_SENTENCES)} function is used to create paragraphs with a specified number of sentences (of random length), and the functions specific to each remaining entity type are used to create an arbitrary number of entities to classify: \textit{ipv4(), ipv4\_private(), ipv4\_public()}, and \textit{ipv6()} for IP addresses, \textit{iban()} for IBAN numbers, and \textit{bitcoin\_address(), ethereum\_address(), litecoin\_address()}, and \textit{polygon\_address()} for blockchain addresses.

Finally, to create the artificial dataset, a random word from the generated paragraph is selected, replaced with one of the above entities, and a label is generated with the entity type, the start and end index of the substitution, and the text of the substituted entity. This process is repeated until there are no more entities to substitute.

\subsection{Text Anonymization with Microsoft Presidio}
Once entities have been located and identified in the text, it is necessary to format the predictions so they can be anonymized with the module offered by \textit{Presidio}. In the case of the identification module of the same tool, this conversion is immediate, requiring only that the output be instantiated under the \textit{RecognizerResult} class. However, for the \textit{fine-tuned} BERT model, it is necessary to extract the located text, its start and end positions, and the confidence score with which the entity is categorized, and convert them to the same class.

Once the outputs from the previous phase have been formatted, it is necessary to instruct the tool on which anonymization technique to use (redaction, masking, encryption, or substitution). Substitution will be used to maintain consistency with the procedure developed by the project partners, allowing for a fair comparison. The substitution method employed will consist of the \textit{class}-\textit{hash} string.

\section{Discussion of Results}
The implementation of information collection from the \textit{Telegram} social network and the transcription of audio messages to text allows for an increase in the number of original intelligence sources from four---text and images on \textit{dark web} and \textit{shallow clear web} pages---to nine, by adding audio to all sources and image and text to \textit{Telegram}. Once the sources have been identified, the results obtained for each module are presented below.

\subsection{Audio Message Transcription}
Table \ref{tab:audio_results} summarizes the WER and execution time for each model, with and without signal preprocessing.

\begin{table}[htb]
	\centering
	\caption{Audio transcription results: WER and execution time}
	\label{tab:audio_results}
	\begin{tabular}{l c c c}
		\hline
		\hline
		Model & WER (raw) & WER (filtered) & Time (s) \\
		\hline
		Parakeet & \textbf{0.16} & \textbf{0.16} & 7215.63 \\
		Wav2Vec  & 0.21          & 0.24           & \textbf{111.93} \\
		Whisper  & 0.37          & 0.40           & 2100.89 \\
		\hline
		\hline
	\end{tabular}
\end{table}

\textit{Parakeet} achieves the best accuracy (WER 0.16), remaining constant regardless of the filter applied. Notably, signal preprocessing is counterproductive for both \textit{Whisper} and \textit{Wav2Vec}, increasing their error rates, which suggests that these models are compromised by external preprocessing. In terms of execution time, \textit{Wav2Vec} is 64$\times$ faster than \textit{Parakeet}, representing a critical trade-off: \textit{Parakeet} is optimal when accuracy is the priority, while \textit{Wav2Vec} is the best option for real-time applications requiring a balance of speed and accuracy (close to 75\% correct).

\subsection{Named Entity Identification}
Table \ref{tab:ner_presidio} shows the \textit{f1 score} and execution time achieved by \textit{Microsoft Presidio} under two hardware configurations.

\begin{table}[htb]
	\centering
	\caption{Microsoft Presidio NER results}
	\label{tab:ner_presidio}
	\begin{tabular}{l c c}
		\hline
		\hline
		Setup & Samples & F1 Score \\
		\hline
		Non-dedicated machine & 100  & 0.74 \\
		HPC cluster            & 225  & 0.79 \\
		\hline
		\hline
	\end{tabular}
\end{table}

While \textit{Presidio} is easy to deploy in production environments, it does not show significant accuracy improvements when more resources are allocated, and execution time grows exponentially beyond a modest sample count, making it not the best solution for the identification phase.

Table \ref{tab:ner_bert} summarizes the BERT-based model results across the three experimental stages.

\begin{table}[htb]
	\centering
	\caption{BERT-based NER results across experimental stages}
	\label{tab:ner_bert}
	\begin{tabular}{l c}
		\hline
		\hline
		Stage (Dataset) & F1 Score \\
		\hline
		Feasibility study (CoNLL-2003) & 0.94 \\
		Fine-tune, train (OntoNotes5) & 0.97 \\
		Fine-tune, validation (ECHR free-form texts) & 0.93 \\
		Cybercrime fine-tune, overall (PII datasets) & 0.96 \\
		Cybercrime fine-tune, minority classes (PII datasets) & 0.53 \\
		\hline
		\hline
	\end{tabular}
\end{table}

The feasibility study (F1 = 0.94) closely matches the 0.95 reported by D. Asimopoulos \textit{et al.} \cite{BenchmarkModels}, validating the choice of base model. The first fine-tuning achieves F1 = 0.97 in training and 0.93 on free-form ECHR texts, demonstrating robustness in unstructured contexts despite texts exceeding the 512-token limit. The cybercrime-specific fine-tuning reaches an overall F1 of 0.96, although performance on underrepresented classes (PASSPORT, IDCARDNUM) drops to 0.53 due to severe class imbalance and high variability in entity formats across countries.

\subsection{Named Entity Anonymization}
Table \ref{tab:anon_results} presents the anonymization metrics for the three evaluated methods.

\begin{table}[htb]
	\centering
	\caption{Anonymization metrics per method}
	\label{tab:anon_results}
	\begin{tabular}{l c c c c c}
		\hline
		\hline
		Method & Info. Loss & Cons. & Coll. & Error Rate & Avg. Corr. \\
		\hline
		Transformer & $-$0.16 & \textbf{1.00} & \textbf{1.00} & \textbf{0.00} & \textbf{0.24} \\
		Presidio    & $-$0.17 & \textbf{1.00} & \textbf{1.00} & \textbf{0.00} & 0.18 \\
		In-line     & 0.26 & 0.00          & 1.00          & 0.50          & 0.05 \\
		\hline
		\hline
	\end{tabular}
\end{table}

Both the \textit{Transformer} and \textit{Presidio} methods achieve perfect consistency and zero error rates, while the negative information loss values indicate that both methods introduce new random tokens during anonymization. The \textit{In-line} method presents a positive information loss (net loss of original information), zero consistency, and an error rate of 0.50, making it inadequate for applications requiring robust and reproducible anonymization. It should be noted that the optimal target values for these metrics may vary depending on the specific application requirements; therefore, the choice of method should always be informed by the context of use.

\section{Selection of the Best Method}
Based on the experimental results, the following integrated \textit{pipeline} is proposed for generating the highest-quality dataset, intended for subsequent behavior analysis and visualization phases:
\begin{enumerate}
	\item \textbf{Intelligence source acquisition}: Web \textit{crawling} tools for the automatic collection of invitation links to messaging groups and channels, over which subsequent search and monitoring of illegal content and possible access to \textit{dark markets} is conducted. Additionally, the OCR mechanism developed by the project partners is employed for text extraction from captured images, thus constituting the capture infrastructure for unstructured sources.
	
	\item \textbf{Audio transcriber}: \textit{Wav2Vec} model, selected for its optimal balance between accuracy and processing speed, suitable for real-time transcription of audio sources.
	
	\item \textbf{Cybercrime entity identifier}: BERT model \textit{fine-tuned} on domain-specific entities (financial accounts, personal identifiers, email addresses, phone numbers, and locations), capable of establishing contextual relationships between entities.
	
	\item \textbf{Anonymizer}: \textit{Transformer}-based system with a \textit{placeholder} substitution method (\textit{class-hash} format), ensuring consistency, traceability, and absence of collisions in the processing of sensitive data.
\end{enumerate}

This configuration maximizes the quantity and quality of processed data, minimizes processing times and error rates, and thus generates the optimal dataset for the subsequent phases of the project.

\section{Conclusions and Future Work}
This work has successfully developed a series of dedicated modules for the acquisition, processing, and anonymization of unstructured intelligence sources in the cybercrime domain, significantly enhancing the analytical capabilities of the technology center. The proposed pipeline expands the number of available intelligence sources from four to nine, incorporating audio transcription and \textit{Telegram} data collection alongside the existing infrastructure. The \textit{Wav2Vec} model emerges as the most balanced transcription solution, offering near-real-time performance at a fraction of the cost in processing time compared to higher-accuracy alternatives. The BERT-based NER model, fine-tuned on cybercrime-specific entities, achieves competitive F1 scores in realistic unstructured scenarios, outperforming \textit{Microsoft Presidio} in both accuracy and scalability. Finally, both the \textit{Transformer}-based and \textit{Presidio} anonymization methods substantially outperform the center's current \textit{In-line} approach, providing consistent, collision-free substitutions with zero error rates.

Given that the project's approach has been to research and develop each module separately, several lines of future work are presented.

First, it would be valuable to integrate all components into a single complete workflow capable of acquiring information sources, analyzing them, and storing the anonymized analysis results.

In the case of content extraction, several lines of work are proposed. First, conduct a more in-depth study of the regulations of different messaging platforms, social networks, and \textit{dark nets} (which may have different terms of use) to integrate more intelligence sources into the information extraction module while complying with their respective terms of service. Second, extend the search for invitation links to private resources through other search engines or \textit{shallow clear web} content indexers, via their APIs or tools. Third, use an image captioning system beyond the OCR currently provided by the \textit{SafeHorizon} consortium to generate more detailed knowledge graphs.

Finally, as mentioned in the previous section, challenges persist related to class imbalance in the datasets used and the absence of entities such as IP addresses, IBANs, and \textit{blockchain} addresses, whose integration---as already proposed but not yet implemented---can be accomplished, for example, through synthetic generation using the \textit{Faker} library to enhance the system's robustness across all entity types relevant to cybercrime analysis.

%%% ACKNOWLEDGEMENT
\section*{Acknowledgements}
This work was partially supported by the European Commission under the Horizon Europe Programme, as part of the project SAFEHORIZON (Grant Agreement No. 101168562). The content of this article does not reflect the official opinion of the European Union. Responsibility for the information and views expressed therein lies entirely with the authors.

%%% References
\bibliographystyle{IEEEtran}
\bibliography{bibliografia}

@misc{Pilan2025TAB,
	author       = {I. Pil{\'a}n and P. Lison},
	title        = {Text Anonymization Benchmark (TAB) v1.0},
	howpublished = {Hugging Face Datasets},
	year         = {2025},
	month        = apr,
	note         = {[Online]. Available: \url{https://huggingface.co/datasets/ildpil/text-anonymization-benchmark} (Last accessed: June 23, 2025)}
}

@inproceedings{tner-ontonotes5,
	title = "{O}nto{N}otes: The 90{\%} Solution",
	author = "Hovy, Eduard  and
	Marcus, Mitchell  and
	Palmer, Martha  and
	Ramshaw, Lance  and
	Weischedel, Ralph",
	booktitle = "Proceedings of the Human Language Technology Conference of the {NAACL}, Companion Volume: Short Papers",
	month = jun,
	year = "2006",
	address = "New York City, USA",
	publisher = "Association for Computational Linguistics",
	url = "https://aclanthology.org/N06-2015",
	pages = "57--60",
}

@inproceedings{conll-2003-ds,
	title = "Introduction to the {C}o{NLL}-2003 Shared Task: Language-Independent Named Entity Recognition",
	author = "Tjong Kim Sang, Erik F.  and
	De Meulder, Fien",
	booktitle = "Proceedings of the Seventh Conference on Natural Language Learning at {HLT}-{NAACL} 2003",
	year = "2003",
	url = "https://www.aclweb.org/anthology/W03-0419",
	pages = "142--147",
}

@misc{espnet-yodas-granary-ds,
	title={Granary: Speech Recognition and Translation Dataset in 25 European Languages}, 
	author={Nithin Rao Koluguri and Monica Sekoyan and George Zelenfroynd and Sasha Meister and Shuoyang Ding and Sofia Kostandian and He Huang and Nikolay Karpov and Jagadeesh Balam and Vitaly Lavrukhin and Yifan Peng and Sara Papi and Marco Gaido and Alessio Brutti and Boris Ginsburg},
	year={2025},
	eprint={2505.13404},
	archivePrefix={arXiv},
	primaryClass={cs.CL},
	url={https://arxiv.org/abs/2505.13404}, 
}

@misc{Fliegner2023BERTConll03,
	author       = {D. Fliegner and T. Khatri and F. Strobel and R. Krestel},
	title        = {bert-large-cased-finetuned-conll03-english},
	howpublished = {Hugging Face Model Hub, dbmdz},
	year         = {2023},
	note         = {[Online]. Available: \url{https://huggingface.co/dbmdz/bert-large-cased-finetuned-conll03-english} (Last accessed: June 23, 2025)}
}

@misc{GoogleBertBaseCased2025,
	author       = {{Google}},
	title        = {bert-base-cased},
	year         = {2025},
	howpublished = {Hugging Face Model Hub},
	url          = {https://huggingface.co/google-bert/bert-base-cased},
	note         = {Accedido: 28-ago-2025}
}

@inproceedings{BenchmarkModels,
	author={Asimopoulos, Dimitris and Siniosoglou, Ilias and Argyriou, Vasileios and Karamitsou, Thomai and Fountoukidis, Eleftherios and Goudos, Sotirios K. and Moscholios, Ioannis D. and Psannis, Konstantinos E. and Sarigiannidis, Panagiotis},
	booktitle={2024 13th International Conference on Modern Circuits and Systems Technologies (MOCAST)}, 
	title={Benchmarking Advanced Text Anonymisation Methods: A Comparative Study on Novel and Traditional Approaches}, 
	year={2024},
	volume={},
	number={},
	pages={1-6},
	doi={10.1109/MOCAST61810.2024.10615642}
}

@article{BalancingWithPresidio,
	author = {Patchipala, Surya},
	year = {2023},
	month = {04},
	pages = {13},
	title = {Data Anonymization in AI and ML Engineering: Balancing Privacy and Model Performance Using Presidio},
	volume = {Volume 6},
	journal = {IRE Journals}
}

@incollection{Hassan2018NER,
	author    = {F. Hassan and J. Domingo-Ferrer and J. Soria-Comas},
	title     = {Anonymization of Unstructured Data via Named-Entity Recognition},
	booktitle = {Modeling Decisions for Artificial Intelligence (MDAI 2018)},
	editor    = {V. Torra and Y. Narukawa and I. Aguil{\'o} and M. Gonz{\'a}lez-Hidalgo},
	series    = {Lecture Notes in Computer Science},
	volume    = {11144},
	pages     = {313--324},
	publisher = {Springer},
	address   = {Cham},
	year      = {2018},
	doi       = {10.1007/978-3-030-00202-2_24}
}

@inproceedings{Hassan2019AutoAnon,
	author    = {F. Hassan and D. S{\'a}nchez and J. Soria-Comas and J. Domingo-Ferrer},
	title     = {Automatic Anonymization of Textual Documents: Detecting Sensitive Information via Word Embeddings},
	booktitle = {2019 18th IEEE International Conference on Trust, Security and Privacy in Computing and Communications / 13th IEEE International Conference on Big Data Science and Engineering (TrustCom/BigDataSE)},
	year      = {2019},
	pages     = {358--365},
	address   = {Rotorua, New Zealand},
	doi       = {10.1109/TrustCom/BigDataSE.2019.00055}
}

@inproceedings{Siniosoglou2024TextAnon,
	author    = {I. Siniosoglou and others},
	title     = {Enhancing Text Anonymisation: A Study on {CRF}, {LSTM}, and {ELMo} for Advanced Entity Recognition},
	booktitle = {2024 Panhellenic Conference on Electronics \& Telecommunications (PACET)},
	year      = {2024},
	pages     = {1--6},
	address   = {Thessaloniki, Greece},
	doi       = {10.1109/PACET60398.2024.10497084}
}

@inproceedings{AsimoEfficacyAI,
	author    = {D. Asimopoulos and others},
	title     = {Evaluating the Efficacy of {AI} Techniques in Textual Anonymization: A Comparative Study},
	booktitle = {2024 7th International Balkan Conference on Communications and Networking (BalkanCom)},
	year      = {2024},
	pages     = {242--246},
	address   = {Ljubljana, Slovenia},
	doi       = {10.1109/BalkanCom61808.2024.10557182}
}

@article{zaratiana2023gliner,
	title={Gliner: Generalist model for named entity recognition using bidirectional transformer},
	author={Zaratiana, Urchade and Tomeh, Nadi and Holat, Pierre and Charnois, Thierry},
	journal={arXiv preprint arXiv:2311.08526},
	year={2023}
}

@misc{UE2016Reglamento679,
	author       = {{Parlamento Europeo y Consejo de la Unión Europea}},
	title        = {Reglamento (UE) 2016/679 del Parlamento Europeo y del Consejo, de 27 de abril de 2016, relativo a la protección de las personas físicas en lo que respecta al tratamiento de datos personales y a la libre circulación de estos datos},
	year         = {2016},
	howpublished = {Diario Oficial de la Unión Europea},
	url          = {https://www.boe.es/doue/2016/119/L00001-00088.pdf},
	note         = {Accedido: 23-jun-2025}
}

@misc{ley_organica_10_1995,
	title = {{Ley Orgánica 10/1995, de 23 de noviembre, del Código Penal}},
	author = {{Jefatura del Estado}},
	year = {1995},
	month = nov,
	day = {23},
	url = {https://www.boe.es/eli/es/lo/1995/11/23/10/con},
	note = {BOE-A-1995-25444}
}

@article{Aben2025WhisperAccessibility,
	author       = {Arypzhan Aben and Gulnur Kazbekova and Zhuldyz Ismagulova and Gulmira Ibrayeva},
	title        = {Audio-to-Text Translation for the Hard of Hearing: A Whisper Model-Based Study},
	journal      = {Scientific Journal of Astana IT University},
	year         = {2025},
	pages        = {24--36},
	doi          = {10.37943/22snok5872},
	publisher    = {Astana IT University}
}

@misc{Marangon2023WER,
	author       = {Johni Douglas Marangon},
	title        = {How to calculate the Word Error Rate in Python},
	year         = {2023},
	month        = nov,
	url          = {https://medium.com/@johnidouglasmarangon/how-to-calculate-the-word-error-rate-in-python-ce0751a46052},
	note         = {Accedido: 23-jun-2025}
}

@misc{Tigerschiold2022AccuracyMetrics,
	author       = {Ted Tigerschiold},
	title        = {What is Accuracy, Precision, Recall and F1 Score?},
	year         = {2022},
	month        = nov,
	howpublished = {Labelf Blog},
	url          = {https://www.labelf.ai/blog/what-is-accuracy-precision-recall-and-f1-score},
	note         = {Accedido: 23-jun-2025}
}

@incollection{KARACA2022231,
	title = {Chapter 14 - Shannon entropy-based complexity quantification of nonlinear stochastic process: diagnostic and predictive spatiotemporal uncertainty of multiple sclerosis subgroups},
	editor = {Yeliz Karaca and Dumitru Baleanu and Yu-Dong Zhang and Osvaldo Gervasi and Majaz Moonis},
	booktitle = {Multi-Chaos, Fractal and Multi-Fractional Artificial Intelligence of Different Complex Systems},
	publisher = {Academic Press},
	pages = {231-245},
	year = {2022},
	isbn = {978-0-323-90032-4},
	doi = {https://doi.org/10.1016/B978-0-323-90032-4.00018-3},
	url = {https://www.sciencedirect.com/science/article/pii/B9780323900324000183},
	author = {Yeliz Karaca and Majaz Moonis}
}

@mastersthesis{DominguezPrieto2023Levenshtein,
	author       = {Alexandre Dom{\'i}nguez Prieto},
	title        = {Distancia de Levenshtein como clasificador de textos},
	school       = {Universidade de Santiago de Compostela},
	year         = {2023},
	type         = {Proyecto de Fin de Máster},
	address      = {Santiago de Compostela, España},
	month        = feb,
	note         = {Directores: Jose Ameijeiras Alonso y Mar{\'i}a Jos{\'e} Ginzo Villamayor. Lectura: 16-feb-2023 (online)},
	url          = {http://eio.usc.es/pub/mte/descargas/ProyectosFinMaster/Proyecto_2142.pdf}
}

@article{samarati1998protecting,
	title={Protecting privacy when disclosing information: k-anonymity and its enforcement through generalization and suppression},
	author={Samarati, Pierangela and Sweeney, Latanya},
	year={1998},
	journal={EPIC, Electronic Privacy Information Center},
	publisher={technical report, SRI International}
}

@article{LdiversityPaper,
	author = {Machanavajjhala, Ashwin and Kifer, Daniel and Gehrke, Johannes and Venkitasubramaniam, Muthuramakrishnan},
	title = {L-diversity: Privacy beyond k-anonymity},
	year = {2007},
	issue_date = {March 2007},
	publisher = {Association for Computing Machinery},
	address = {New York, NY, USA},
	volume = {1},
	number = {1},
	issn = {1556-4681},
	url = {https://doi.org/10.1145/1217299.1217302},
	doi = {10.1145/1217299.1217302},
	journal = {ACM Trans. Knowl. Discov. Data},
	month = mar,
	pages = {3–es},
	numpages = {52}
}

@INPROCEEDINGS{tclosenessPaper,
	author={Li, Ninghui and Li, Tiancheng and Venkatasubramanian, Suresh},
	booktitle={2007 IEEE 23rd International Conference on Data Engineering}, 
	title={t-Closeness: Privacy Beyond k-Anonymity and l-Diversity}, 
	year={2007},
	volume={},
	number={},
	pages={106-115},
	doi={10.1109/ICDE.2007.367856}}

@inproceedings{nergiz2007hiding,
	title={Hiding the presence of individuals from shared databases},
	author={Nergiz, Mehmet Ercan and Atzori, Maurizio and Clifton, Chris},
	booktitle={Proceedings of the 2007 ACM SIGMOD international conference on Management of data},
	pages={665--676},
	year={2007}
}

@misc{zaratiana2023glinergeneralistmodelnamed,
	title={GLiNER: Generalist Model for Named Entity Recognition using Bidirectional Transformer}, 
	author={Urchade Zaratiana and Nadi Tomeh and Pierre Holat and Thierry Charnois},
	year={2023},
	eprint={2311.08526},
	archivePrefix={arXiv},
	primaryClass={cs.CL},
	url={https://arxiv.org/abs/2311.08526}, 
}
\end{document}